\documentclass{article}
\usepackage{spconf,amsmath,graphicx}
\usepackage{multirow}
\usepackage{mathptmx}
\usepackage[medium, compact ]{titlesec}

\title{axonal delay as a short-term memory for Feed forward \\Deep Spiking Neural Networks}

\name{Pengfei Sun$^{1}$ \qquad Longwei Zhu$^{2*}$\thanks{$^{*}$Corresponding Author}\thanks{
This research received funding from the Flemish Government under the "Onderzoeksprogramma Artificiele Intelligentie (AI) Vlaanderen" and Programmatic grant no. A1687b0033 from the Singapore Governments Research, Innovation and Enterprise 2020 plan (Advanced Manufacturing and Engineering domain).} \qquad Dick Botteldooren$^{1*}$} 

\address{$^{1}$ Ghent University, Belgium \\
    $^{2}$ Institute for Infocomm Research, A*STAR, Singapore}

\begin{document}
%
\maketitle
\begin{abstract}
   
The information of spiking neural networks (SNNs) are propagated between the adjacent biological neuron by spikes, which provides a computing paradigm with the promise of simulating the human brain. Recent studies have found that the time delay of neurons plays an important role in the learning process. Therefore, configuring the precise timing of the spike is a promising direction for understanding and improving the transmission process of temporal information in SNNs. However, most of the existing learning methods for spiking neurons are focusing on the adjustment of synaptic weight, while very few research has been working on axonal delay. In this paper, we verify the effectiveness of integrating time delay into supervised learning and propose a module that modulates the axonal delay through short-term memory. To this end, a rectified axonal delay (RAD) module is integrated with the spiking model to align the spike timing and thus improve the characterization learning ability of temporal features. Experiments on three neuromorphic benchmark datasets : NMNIST,  DVS Gesture and N-TIDIGITS18 show that the proposed method achieves the  state-of-the-art  performance while using the fewest parameters.

\end{abstract}


\section{Introduction}

Spiking neural networks, which are composed of biologically plausible spiking neurons, have been proven to be robust on several tasks for both unsupervised\cite{gerstner1996neuronal,song2000competitive,zhang2018highly} and supervised learning\cite{bohte2002error,7968359}. Not only the visual cognitive tasks, recent researches also showed the enormous potential of spiking neural networks on acoustical tasks\cite{pan2018event,pan2020efficient}. The spiking neuron, which generates spikes when its membrane potential exceeds the threshold usually relies on spiking events in time to propagate the information forward. Thus, SNNs are inherently suited for problems of temporal nature. With the fiery and tremendous growth of artificial intelligence in various research areas, 
a series of multi-layer spiking neural networks have also been rapidly developed for different learning tasks\cite{Shrestha2018,xu2013supervised,shrestha2016event,kheradpisheh2020temporal,zhang2021rectified}.

However, the randomness in the spiking character of an SNN itself dictates its imperfection. With the inexplicable nature of the inner activation of the multi-layer network, the assignment of temporal credit is a huge problem. Several delay-based learning paradigms have been proposed to solve such discrepancies problem. Knoblauch and Sommer \cite{knoblauch2004spike} discussed short and long delays effects in Spike Timing Dependent Plasticity (STDP) rule, while Zhang et al. \cite{zhang2020supervised}   proposed that synaptic delay can be used to learn the precise spiking time by affecting the membrane potential in a multi-layer spiking neural network. Nevertheless, these methods completely ignore the possibility of modifying axonal delay in deep spiking neural networks. 
Neurophysiology evidence suggests that axonal delay modulation can occur as a short-term memory
during the learning process, greatly affecting learning performance. Researches also show that the axonal delays could be the fundamental of the different response characteristics of different neuronal populations\cite{carr1988axonal,stoelzel2017axonal}.  

In this article, we propose modulation of axonal delay in the deep spiking neural networks based on the supervised learning methods. We describe the updating rule of the rectified axonal delay(RAD) module and show how this module behaves in the spiking neuron. The comprehensive experiments show that deep SNNs with our proposed RAD module could significantly outperform the benchmarks in terms of accuracy and model size.

\label{sec:intro}

\section{PROPOSED METHOD}
\label{sec:format}
In this work, we propose a rectified axonal delay (RAD) module to align the spike firing timing during the pulse transmission along the axon. This module is implemented on top of the spiking neuron model and is expected to improve the representation power especially for time-critical identification tasks. 

\subsection{Spiking Neuron Model}
The spiking neuron is the computational unit of SNNs that communicates with spikes and maintains an interval membrane potential over time. In this paper, 

we adapt the spike response model (SRM) as the neuron model and formulate our method based on SLAYER-PyTorch \cite{Shrestha2018}, which is an effective and powerful training framework.    

The sub-threshold membrane potential of the SRM neuron can be described as follows\\
\begin{equation}
          u^{l}_{i}(t) = \sum_{j}(W^{l-1}_{ij}(\epsilon * s^{l-1}_{j})(t) + (\nu * s^{l}_{j} )(t)) 
\end{equation}

Where $u^{l}_{i}(t)$ is the membrane potential of neuron $i$ in layer $l$ at time $t$ and $s^{l-1}_{j}$ are the incoming spikes. $\epsilon(t)$ and  $\nu(t)$ denote spike response kernel and refractory kernel respectively. The incoming spikes are converted into a response signal by convolving incoming spikes with a spike response kernel. According to Eq.1, the neuron's state is updated by the sum of the postsynaptic potential (PSP) and refractory response, where the PSP is the weighted sum of the response signal from other neurons and the refractory phase describes the brief period before the neuron regains its capacity to make a second response.
The neuron generates an output spike when the $u_{i}(t)$ surpasses the predefined threshold $\theta_{u}$ and transmits the spikes to the subsequent neurons along the axon. This generation process can be formulated by a Heaviside step function $\Theta$ as follows
\begin{equation}
          s^{l}_{i}({t}) = \Theta({u^{l}_{i}(t) - \theta_{u}}) \
\end{equation}
 
\subsection{Rectified Axonal Delay Module}

The axonal delay module of each neuron simulates the delay in the transmission process. We denote $N$ is the number of neurons at layer $l$, thus, the spike train $s^{l}(t)$ can be represented as follows
\begin{equation}
           s^{l}(t) = \{s^{l}_{1},...,s^{l}_{N}\}\\
\end{equation}
To speed up the training and the response of the networks, we limit the time delay of each neuron to a reasonable range. 
\begin{equation}
           \hat s_{d}^{l}(\hat{t}) = \delta(t-\hat{d^{l}}) * s^{l}(t)
\end{equation}
\begin{equation}
          \hat{d}^{l} = \begin{cases}
          0 &  \text{$d < 0$}\\
          d &  \text{$0 \leq d \leq\theta_{d}$}\\
          \theta_{d} &  \text{$\theta_{d} < d $}
          \end{cases}
\end{equation}
The $\theta_{d}$ refers to the limit delay of the spiking neuron and $\hat s_{d}^{l}(\hat{t})$ is the shifted spike trains. 

During the training, $\widetilde{ s}^{l}({t})$ is the desired spike trains for all output neurons at time $t$. Then the loss function $L$ is usually defined as
\begin{equation}
           L = ( \sum_{t=0}^{T       }\widetilde{ s}^{l}({t}) - \sum_{t=0}^{T       }s^{l}({t}))^2  
\end{equation}

The temporal dimension is discretized with the sampling time $T_s$ such that $t = nT_s$. With $N_s$ denoting the total number of samples, the observation time becomes $T=(N_s-1)T_s$. Taking into account the temporal dependency, the gradient term of synaptic weight is the same as Shrestha et al. \cite{Shrestha2018} except there is a delay shift of spike. For the axonal delay of layer $l$, the gradient term is given by
\begin{equation}          \bigtriangledown      \hat d^{l} = T_s         \sum_{n=0}^{N_s       }\frac{\partial L[n]}{\partial          \hat d^{l}}   \
\end{equation} 
Here $L[n]$ is the loss at time instance $n$. Using the chain rule and understanding the fact that the loss $L[n]$ is dependent on all previous values of shifted spike trains. we get 
\begin{equation}          \bigtriangledown      \hat d^{l} =  T_s         \sum_{n=0}^{N_s       }\sum_{m=0}^{n}        \frac{\partial         \hat s_{d}^{l}[m]}{\partial          \hat d^{l}}  
       \frac{\partial L[n]}{\partial \hat s_{d}^{l}[m]} \
\end{equation} 
We use finite difference approximation $\frac{\hat s_{d}^{l}[m]- \hat s_{d}^{l}[m-1]}{T_s}$ to numerically estimate the gradient term $\frac{\partial \hat s_{d}^{l}[m]}{\partial \hat d^{l}}$, where $m-1$ refers to the previous time step. After formulating the forward and backward propagation, the axonal delay and synaptic weight are updated through the gradient descent algorithm.

\section{EXPERIMENTS AND RESULTS}
\label{sec:pagestyle}
In this session, we conduct a series of experiments to validate the performance of our proposed method. In this paper, we will follow the similar notation as Shrestha et al. \cite{Shrestha2018} to define our network architecture. The layer is separated by the $-$, a convolution layer with $x$ channels and $y$ filters are represented by the $_xc_y$, an aggregation (pooling)  layer with $y$ filters is represented by $a_y$. The input signal is expressed as $H \times W\times C$, where $H$ and $W$ are the spatial dimensions and $C$ is the input channel.  

\subsection{IMPLEMENTATION DETAILS}
In our experiments, we use the SLAYER-PyTorch as our training framework and implement the rectified axonal delay (RAD) module on top of it. Each network and RAD module are trained with an identical optimizer. We use the number of spikes generated from the last layer as the loss measurement and specify the desired spikes for different tasks respectively. We use the response kernel $\epsilon(t) = \frac{t}{\tau_s}\exp(1-\frac{t}{\tau_s})\Theta(t)$ and refractory kernel
$\nu(t) = -2\theta_{u}\,\frac{t}{\tau_r}\exp(1-\frac{t}{\tau_r})\Theta(t)$. Here, ${\tau_s}$ and ${\tau_r}$ are the time constant of the kernels. The simulation step time ${T_s}$ is set as 1 ${ms}$. Table  \ref{tbl:param} lists all the hyper-parameters we used in our three examples. All the experiments are run for 5 independent trials and we report the average performance and deviation for a fair comparison. Our codes and the detailed configuration are made publicly
available\footnote{The codes is available at:https://github.com/bamsumit/slayerPytorch}. 

\begin{table}
\small
	\centering
	\caption{
Detailed hyper-parameter settings for different datasets}
	\label{tbl:param}
	\begin{tabular}{clrccc}
		\cline{1-5}
		\multicolumn{1}{c}{\bf Dataset}& \multicolumn{1}{c}{\bf $\tau_s$} & \multicolumn{1}{c}{\bf $\tau_r$} & \textbf{$\theta_{d}$}  &\textbf{$\theta_{u}$}
		\\ \hline
	\multirow{1}{*}{NMNIST}
	& 1 	& 1  & 64 &10 \\
	\multirow{1}{*}{{DVS Gesture}}
		& 5 	& 5  & 64&10  \\
\multirow{1}{*}{{N-TDIDIGITS18}}
		&  5		& 5 & 128&10 \\
	\end{tabular}
	\vspace{-0.5cm}
\end{table}

\subsection{DATASETS AND OVERALL RESULTS}

\subsubsection{NMNIST}
The NMNIST\cite{orchard2015converting} dataset contains 60000 training samples and 10000 testing samples originated from the MNIST image and then processed by the Dynamic Vision Sensor(DVS). Each sample has a 300 ms duration and the spatial dimension is $34 \times 34$ pixels. In our experiment, we only use the raw data and don't do any processing to compensate for the saccadic movement. 

\begin{table}
\small
	\centering

	\caption{Comparison with the state-of-the-art in terms of network size and accuracy.}
	\label{tbl:results}
     \setlength{\tabcolsep}{0.5mm}{
	\begin{tabular}{|c|l|r|c|}
		\cline{2-4}
		\multicolumn{1}{c|}{}& \multicolumn{1}{c|}{\bf Method} & \multicolumn{1}{c|}{\bf Params} & \textbf{Accuracy}
		\\ \hline
		\multirow{4}{*}{NMNIST}
		& Tandem learning. \cite{wu2021tandem} & 
		4.63 MB& $99.31\%$ \\
		& Wu et al. \cite{wu2019direct}	& 
		17.67 MB& $99.53\%$ \\
		&Spike-based BP. \cite{fang2021incorporating}& 65.36 MB&\bf99.61\% \\
		&\bf Our method	& 	\bf 2.13 MB & 99.37\% \\
		\hline
		\multirow{5}{*}{{DVSGesture}}
		& TrueNorth \cite{amir2017low}
						& 
						1.99 MB& $91.77 (94.59)\%$ \\

		& DECOLLE \cite{kaiser2020synaptic}	 
		                & 
		                1.25 MB& $95.54\%$ \\
		& Ghosh et al. \cite{ghosh2019spatiotemporal}$^\dagger$ 
		                & 
		                2.12 MB& $95.94\%$ \\
		&Spike-based BP. \cite{fang2021incorporating}& 6.48 MB&\bf97.57\% \\
		& \bf Our method		& \bf 1.06 MB& 96.97\% \\
		\hline
		\multirow{4}{*}{{NTDIDIGITS}}
		& GRU-RNN \cite{anumula2018feature}$^\dagger$
						& 
						0.11 MB& $90.90\%$ \\
		& Phased-LSTM \cite{anumula2018feature}$^\dagger$
						& 
						0.61 MB& $91.25\%$ \\

		& ST-RSBP \cite{zhang2019spike}
						& 
						0.35 MB& $93.63\pm 0.27\%$ \\
		& \bf Our method		&   

		\bf0.08 MB& $\bf94.45\%$  \\
		\hline
		\multicolumn{3}{l}{\footnotesize{$^\dagger$ Non SNN implementation.}}
	\end{tabular}}
	\vspace{-0.5cm}
\end{table}

For this task, the following spiking CNN architecture is used: \verb~34x34x2-16c5-a2-32c3-a2-64c3~ \verb~-512-10~, wherein the pure numbers refer to the number of neurons at each fully-connected layer. As can be seen from Table \ref{tbl:results}, our method is very competitive compared with other benchmarks. The performance is a little lower than the best-reported result\cite{fang2021incorporating}, while the model size is significantly reduced (30X). This kind of small size model will benefit more by combining the mapping technique \cite{gopalakrishnan2020hfnet}.   
\subsubsection{DVS Gesture}
The DVS-Gesture \cite{amir2017low} is the dataset consisting of 29 subjects performing 11 different hand and arm gestures. These gestures are recorded using the DVS camera under 3 illumination conditions. Unlike the NMNIST, this task is not derived from the static image but the real movement of the subject. We use the first 23 subjects for training while the last 6 for testing.  

An SNN with architecture \verb~128x128x2-a4-16c5-a2~
\verb~-32c3-a2-512-11~ was trained in this dataset. To speed up training, we only use randomly selected 300ms long sequences. while for the inference, the first 1.5 $s$ of action video for each class is used to classify the actions. The results are listed in Table \ref{tbl:results}. Our method shows the best accuracy of $96.21\%$ on average and our model size is only 1 $MB$ . 

\subsubsection{N-TDIDIGITS18}
The N-TDIDIGITS18 \cite{anumula2018feature} dataset is the neuromorphic version of TDIDIGITS \cite{leonard1993tidigits}. It contains the 11 spoken digits (“oh,” and the digits “0” to “9”) and 64 response channels. We use the same train-test split of the dataset as \cite{anumula2018feature}.  

The fully connected architecture \verb~64-256-256-11~  is explored. As we can see from Table \ref{tbl:results}, we report the best performance of $94.45\%$ with a mean of $94.19\%$ and a stand deviation of $0.18\%$. Our method significantly outperforms the Non-SNN based methods and other SNN based approaches. It is worth noting that our method can classify with fewer parameters and better performance. Compared with nonspiking networks like Phased Long Short Term Memory (Phased LSTM) \cite{anumula2018feature}, we can achieve $2.94\%$ performance improvement while using $7X$ fewer parameters.   

\subsection{DETAILED ANALYSIS OF THE RESULTS}

The effect of the proposed RAD module introducing the axonal delay function is further analyzed. As shown in Table \ref{tbl:abd}, it can be observed that: (1) For the NMNIST dataset, which is recorded from static images and does not comprise much temporal information, the performance is only slightly better when combining spiking CNN network with the RAD module, and the improvement of adding a delay is not significant. This indicates that precise spike timing is not important in this dataset \cite{iyer2021neuromorphic}. (2) For the event-based video dataset DVS Gesture and neuromorphic audio dataset N-TDIDIGITS18 which are both highly temporal-dependent, the proposed method introduces rapid performance gains with added delay, which demonstrate that the axonal delay contributes to optimizing the spike temporal information that enhances the representation of features. As can be seen from the table, the DVS Gesture classification accuracy is boosted by 0.57\% with the axonal delay. The most considerable performance improvement comes from the speech task, and the best accuracy can reach 94.45\%, which is more than 15\% better. 


During the experiments with axonal delay studies, it is found that after a large amount of repetitive training, the spike occurrence can be over-inhibited. That means a spike might be depressed for too long. As a result, for the speech task from the table \ref{tbl:abd}, when the $\theta_d=+\infty$, the classification performance becomes worse. Thus, it is necessary to constrain the shifting range to prevent such over-inhibition. 

	\begin{figure*}[]
		\centering
	\includegraphics[scale=0.54]{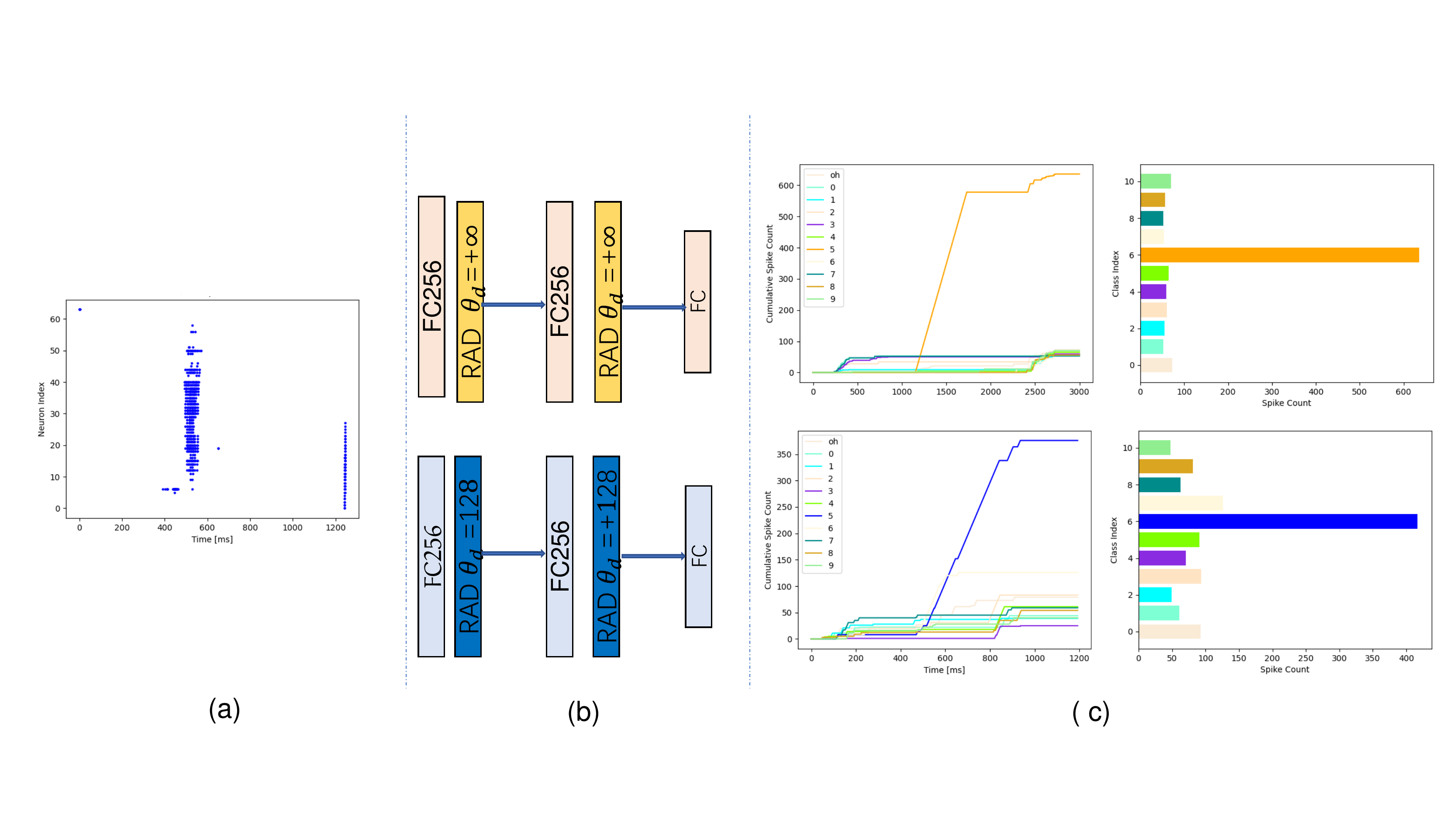}\vspace{-1.5cm}
		\caption{Illustration of flow chart of the proposed method. (a) Spiketrains of input sample `5'. (b) Three fully-connectd layers with two different threshold scheme of delay. (c) Cumulative Spikecount of last layer (left),  Spikecount(right)}
	\label{fig:example}
	\end{figure*}

A simple experiment helps to understand the effectiveness of the RAD module. We take voice `5' in the NTIDIGITS as an input example (Figure 1(a)) and explore the three fully-connected networks with different axonal delay limitations (Figure 1(b)). Figure 1(c) left and right illustrate the cumulative spike count distribution over time and the total spike count for every output neuron respectively. For both cases, the total spike count is used to decide which number was spoken, and both are well classified. While for the system delay, as can be observed from the Figure 1(c) left, the rectified axonal delay could classify the sample as early as 800 $ms$ , while the unlimited axonal delay needs 1500 $ms$ to vote for the true class.

\begin{table}
\small
	\centering
	\caption{Ablation studies for different delay threshold $\theta_{d}$ in the RAD module.}
	\label{tbl:abd}
	\begin{tabular}{clrc}
		\cline{1-4}
		\multicolumn{1}{c}{\bf Dataset}& \multicolumn{1}{c}{\bf $\theta_{d}$} & \multicolumn{1}{c}{\bf Params} & \textbf{Accuracy}
		\\ \hline
		\multirow{2}{*}{NMNIST}
		& 0  &  
		2,126,754& $99.26\pm0.02\%$ \\
		&  \textbf{64}	& 
		2,131,458& $\bf 99.33\pm 0.03\%$  \\
		
		\hline
		\multirow{2}{*}{{DVS Gesture}}
		&  0		& 
		1,060,211& $95.64\pm 0.65\%$ \\
		& \bf 64		&  
		1,060,771& $\bf96.21\pm 0.63\%$ \\
		\hline
		\multirow{2}{*}{{N-TDIDIGITS18}}
		&  0		& 
		85,259& $78.86\pm 0.47\%$ \\
		&  $+\infty$		&  
		85,771& $93.83\pm 0.10\%$ \\		
		& \bf128		&  
		85,771& $\bf94.19\pm 0.18\%$ \\
		\hline

	\end{tabular}
	\vspace{-0.5cm}
\end{table}

\section{CONCLUSIONS}
\label{sec:typestyle}

Spiking neural networks (SNNs) offer the opportunity to include precise timing as part of the solution for classifying objects with a temporal aspect. However, this possibility has been rarely used in its full potential. In this work, we introduce the rectified axonal delay (RAD) as an additional degree of freedom for training that can easily be incorporated into existing SNN frameworks. 
The comprehensive evaluation results show that the proposed RAD module could significantly outperform several other models on two out of the three benchmarks that were selected. Not surprisingly, the new model performs particularly well on problems where timing matters. We believe that using spiking neuron delay to model short-term memory needed to interpret a spoken word or a gesture, explain the performance increase and the reduction of parameters needed. Biological evidence conceptually supports this statement.
 
In addition, such module
can be easily incorporated into the current deep spiking models with very few tunable parameters added.

\section{ACKNOWLEDGEMENTS}
\label{sec:foot}
The authors would express our very great appreciation to Dr. Sumit Bam Shrestha for his valuable and constructive suggestions and technical support during the development of this research work. 

\small
\bibliographystyle{IEEEbib}
\bibliography{refs}

\vfill\pagebreak


\end{document}